\documentclass[12pt]{article}
\usepackage[utf8]{inputenc}
\usepackage{amsmath}
\usepackage{amssymb}

\usepackage[table,xcdraw, dvipsnames]{xcolor}

\usepackage[font=small,labelfont=bf,labelsep=space]{caption}
\usepackage{xspace}
\usepackage[pdftex]{graphicx}
\usepackage[hidelinks]{hyperref}
\hypersetup{
    colorlinks = true,
    linkcolor={gray!90!black},
    citecolor={gray!90!black},
    urlcolor={gray!90!black}
}
\usepackage{authblk}
\usepackage{geometry}
\usepackage{parskip} 
\usepackage[inline]{enumitem}

\usepackage{booktabs}  
\usepackage{amsfonts}
\usepackage{mathtools}
\usepackage{url}
\usepackage{enumitem}
\usepackage{authblk}
\usepackage{comment}
\usepackage{mdframed}
\usepackage{wrapfig}
\usepackage{floatrow}

\newcommand{\eg}{\emph{e.g.}\xspace}
\newcommand{\ie}{\emph{i.e.}\xspace}

\newcommand{\xhdr}[1]{\vspace{1.3mm}\noindent{{\bf #1.}}}
\newcommand{\xhdrd}[1]{\vspace{1.3mm}\noindent{{\bf #1}}}

\usepackage{ntheorem}

\theoremseparator{:} 

\newmdtheoremenv[%
  backgroundcolor=white,
  linecolor=blue!60!black,
  linewidth=2pt,
  topline=true,
  rightline=false,
  skipabove=10pt,
  skipbelow=10pt,
  leftline=false]{ourexample}{Application}
  
\newmdtheoremenv[%
  backgroundcolor=gray!20,
  linecolor=red!60!black,
  linewidth=2pt,
  topline=false,
  rightline=false,
  skipabove=10pt,
  skipbelow=10pt,
  leftline=false]{ourbox}{Formulation}

\newmdtheoremenv[%
  backgroundcolor=gray!20,
  linecolor=red!60!black,
  linewidth=2pt,
  topline=false,
  rightline=false,
  skipabove=10pt,
  skipbelow=10pt,
  leftline=false]{regbox}{Box}

\theoremstyle{nonumberplain}
\newmdtheoremenv[%
  backgroundcolor=gray!20,
  linecolor=red!60!black,
  linewidth=2pt,
  topline=false,
  rightline=false,
  skipabove=10pt,
  skipbelow=10pt,
  leftline=false]{suppregbox}{Box S1}

\definecolor{Gray1}{gray}{0.82}
\definecolor{Gray2}{gray}{0.92}

\pdfmapfile{=Cochineal.map}
\usepackage[p,osf]{cochineal}
\usepackage[T1]{fontenc}

\usepackage[scale=.95,type1]{cabin}
\usepackage[zerostyle=c,scaled=.94]{newtxtt}
\usepackage[cal=boondoxo]{mathalfa}
\usepackage{microtype}

\usepackage{etoolbox}
\apptocmd{\thebibliography}{\raggedright}{}{}
\def\gA{{\mathcal{A}}}
\def\gE{{\mathcal{E}}}
\def\gN{{\mathcal{N}}}
\def\rmA{{\mathbf{A}}}

\def\rmH{{\mathbf{H}}}

\def\rmZ{{\mathbf{Z}}}

\title{Multimodal learning with graphs
}
\author[1,2,$*$]{Yasha Ektefaie}
\author[2,5,$*$]{George Dasoulas}
\author[2,3]{Ayush Noori}
\author[2,6]{\\Maha Farhat}
\author[2,4,5,$\ddag$]{Marinka Zitnik}
\affil[1]{\small Bioinformatics and Integrative Genomics Program, Harvard Medical School, Boston, MA 02115, USA}
\affil[2]{Department of Biomedical Informatics, Harvard  University, Boston, MA 02115, USA}
\affil[3]{Harvard College, Cambridge, MA 02138, USA}
\affil[4]{Broad Institute of MIT and Harvard, Cambridge, MA 02142, USA}
\affil[5]{Harvard Data Science Initiative, Cambridge, MA 02138, USA}
\affil[6]{Division of Pulmonary and Critical Care, Department of Medicine, Massachusetts General Hospital, Boston, MA, USA\vspace{2mm}}
\affil[$\ddag$]{Correspondence: marinka@hms.harvard.edu}
\affil[$*$]{Equal contribution}
\date{}

\begin{document}

\maketitle

\begin{abstract}
\noindent Artificial intelligence for graphs has achieved remarkable success in modeling complex systems, ranging from dynamic networks in biology to interacting particle systems in physics. However, the increasingly heterogeneous graph datasets call for multimodal methods that can combine different inductive biases---the set of assumptions that algorithms use to make predictions for inputs they have not encountered during training. Learning on multimodal datasets presents fundamental challenges because the inductive biases can vary by data modality and graphs might not be explicitly given in the input. To address these challenges, multimodal graph AI methods combine different modalities while leveraging cross-modal dependencies using graphs. Diverse datasets are combined using graphs and fed into sophisticated multimodal architectures, specified as image-intensive, knowledge-grounded and language-intensive models. Using this categorization, we introduce a blueprint for multimodal graph learning, use it to study existing methods and provide guidelines to design new models.
\end{abstract}



\section{Introduction}
Deep learning on graphs has contributed to breakthroughs in biology~\cite{Greener2022,michael2018visible}, chemistry~\cite{moleculenet,pmlr-v70-gilmer17a}, physics~\cite{sanchezgonzalez2018graph,sanchezgonzalez2020learning}, and the social sciences~\cite{liu2020survey}. The predominant use of graph neural networks~\cite{scarselli_gnn} is to learn representations of various graph components---such as nodes, edges, subgraphs, and entire graphs---based on neural message passing strategies. The learned representations are used for downstream tasks, including label prediction via semi-supervised learning~\cite{kipf_gcn}, self-supervised learning~\cite{vgae}, and graph design and generation~\cite{graphgen1,graphgen2}. In most existing applications, datasets explicitly describe graphs in the form of nodes, edges, and additional information representing contextual knowledge, such as node, edge, and graph attributes. 

\begin{figure}[t]
    \centering
    \includegraphics[width=\textwidth]{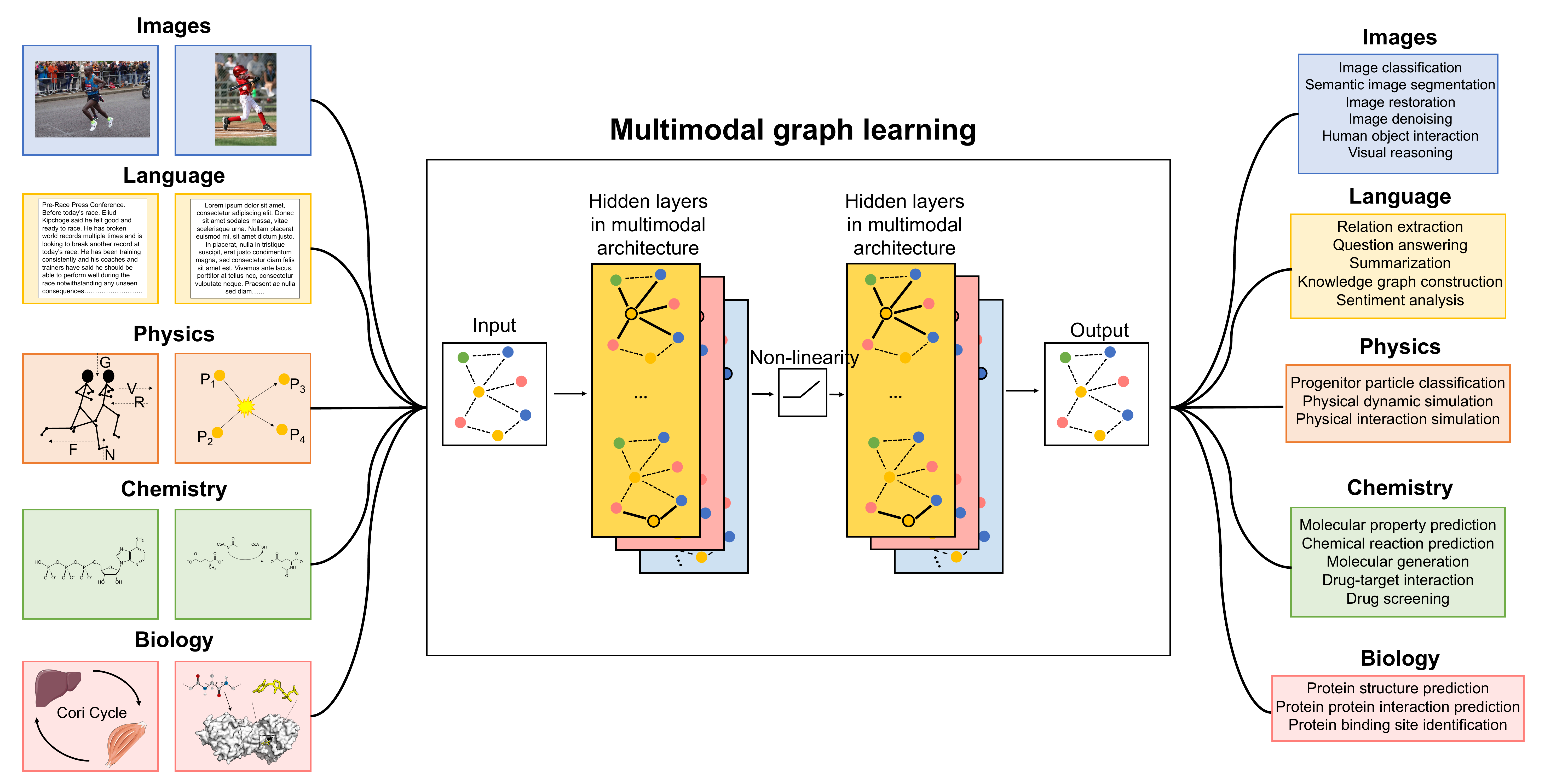}
    \caption{\textbf{Graph-centric multimodal learning}. Shown on the left are the different data modalities. Shown on the right are machine learning tasks for which multimodal graph learning has proved valuable. We introduce the multimodal graph learning (MGL) blueprint that serves as a unifying framework for multimodal graph neural architectures realized through learning systems in computer vision, natural language processing, and natural sciences. }
    \label{fig:mgl_paper_overview}
\end{figure}

Modeling complex systems requires measurements that describe the same objects from different perspectives, at different scales, or through multiple modalities, such as images, sensor readings, language sequences, and compact mathematical statements. Multimodal learning~\cite{multimodalsurveytaxonomoy} studies how such heterogeneous, complex descriptors can be optimized to create learning systems that are broadly generalizable, robust to changes in the underlying data distributions, and can train more with less labeled data. While multimodal learning has been successfully used in settings where unimodal methods fail~\cite{mmm_success_1,mm_success_2,mm_success_3}, it presents several challenges that must be overcome to enable its broad use in AI~\cite{multimodalsurveytaxonomoy, mm_performance_gap}. These challenges include finding representations optimized for machine learning analyses and fusing combined information from various modalities to create predictive models~\cite{transformer_multimodal_survey,Bayoudh2022visionmultimodal, zhang2020multimodalfusion}. These challenges have proven difficult. For example, multimodal methods tend to focus on only a subset of modalities that are most helpful during model training while ignoring modalities that might be informative for model implementation---a pitfall known as \textit{modality collapse}~\cite{modality_collapse}. 
Moreover, in contrast to the frequent assumption that every object must exist in all modalities, the complete set of modalities is rarely available due to limitations of data collection and measurement technologies---a challenge known as \textit{missing modalities}~\cite{missing_modality_SMIL_2021, missing_modality_icml_2022}. Because different modalities can lead to intricate relational dependencies, simple modality fusion cannot fully leverage multimodal datasets~\cite{zitnik2019machine}. Graph learning models such data systems~\cite{jumper2021alphafold, somnath2021multiscale,walters2021} by connecting data points in different modalities as edges in optimally defined graphs~\cite{multigcn, graphfusionetwork,zhang2022graph} and building learning systems for a wide range of tasks~\cite{multigraphcontrastive, mmgl}. 

We introduce a blueprint for multimodal graph learning (MGL). The MGL blueprint provides a framework that can express existing algorithms and help develop new methods for multimodal learning leveraging graphs. This framework allows for learning fused graph representations and studying the aforementioned challenges of modality collapse and missing modalities~\cite{transformer_multimodal_survey, multimodalsurveytaxonomoy}. We apply this formulation across a broad spectrum of domains, ranging from computer vision and language processing to the natural sciences (Figure~\ref{fig:mgl_paper_overview}). We consider image-intensive graphs (IIG) for image and video reasoning (Section~\ref{sec:images}), language-intensive graphs (LIG) for processing natural and biological sequences (Section~\ref{sec:language}), and knowledge-intensive graphs (KIG) used to aid in scientific discovery (Section~\ref{sec:knowledge}).


\section{Graph Neural Networks for Multimodal Learning}\label{sec:multimodal}
Deep learning has created a wide range of fusion approaches for multimodal learning~\cite{multimodaldeep,multimodaldeep2}. For example, recurrent neural network (RNN) and convolutional neural network (CNN) architectures have successfully been combined to fuse sound and image representations in video description problems~\cite{videocaptioning1, videocaptioning2}. More recently, generative models have also proven very accurate for both language-dependent~\cite{multimodal_NLM} and physics-based multimodal data~\cite{physics_multimodal_aaai}.    
Such models are based on an encoder-decoder framework, where in the encoder, the combined architectures are trained simultaneously (each one specialized for a modality), while the decoder aggregates information from individual architectures. When complex relations between modalities produce a network structure, graph neural networks (GNNs, Supplementary Note~1) provide an expressive and flexible strategy to leverage interdependencies in multimodal datasets. 

\begin{figure}[t]
    \centering
    \includegraphics[width=\textwidth]{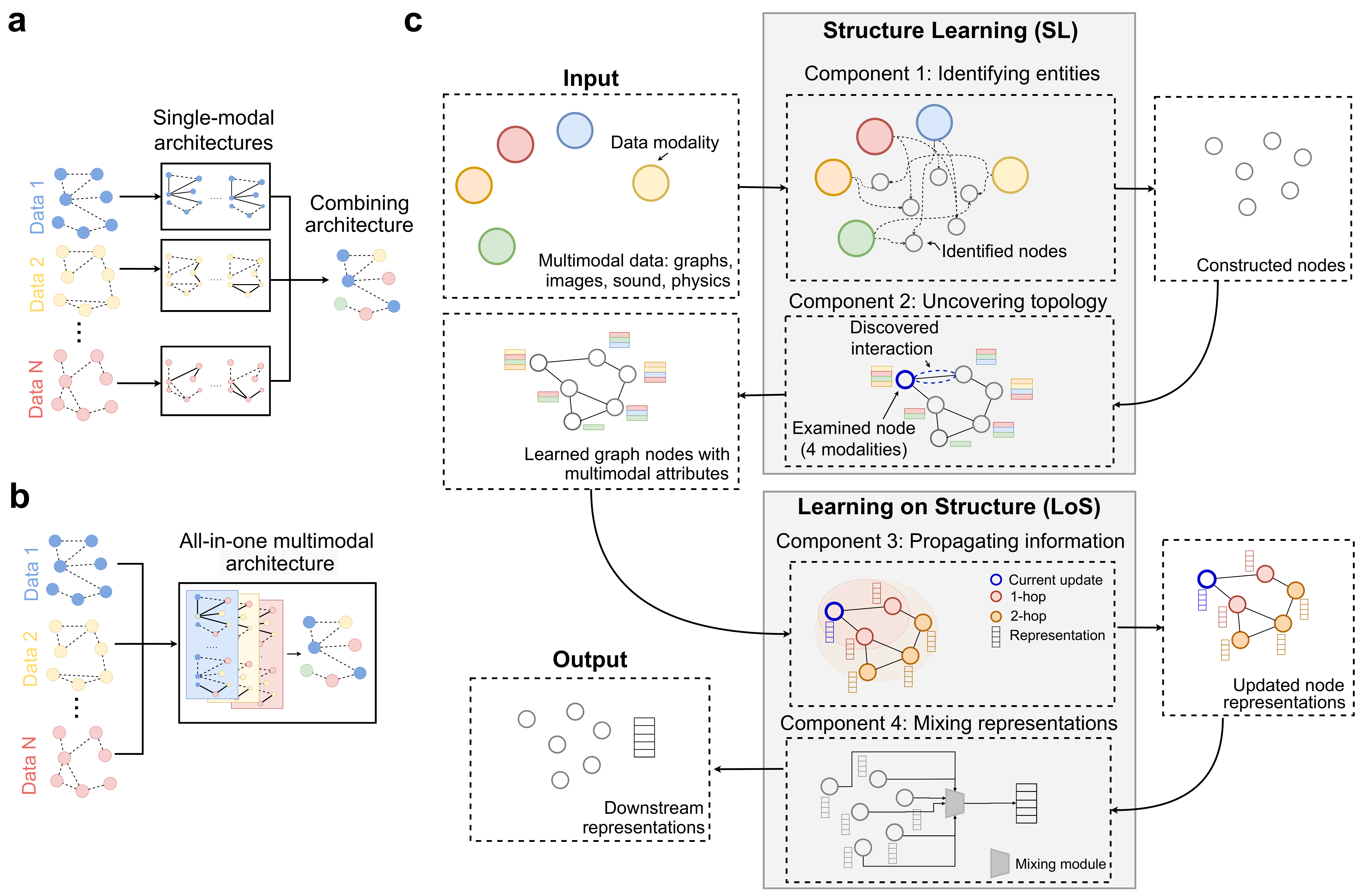}
    \caption{\textbf{Overview of multimodal graph learning (MGL) blueprint}. \textbf{a,} A standard approach to multimodal learning involves combining different unimodal architectures, each optimized for a distinct data modality. \textbf{b,} In contrast, an all-in-one multimodal architecture considers inductive biases specialized for each data modality and optimizes model parameters in an end-to-end manner, enabling expressive data fusion. \textbf{c,} The MGL blueprint comprises four components: identifying entities, uncovering topology, propagating information, and mixing representations. These components are grouped into two phases: structure learning and learning on the structure.}
    
    \label{fig:mgl_blueprint}
\end{figure}

\subsection{Blueprint for Graph-Centric Multimodal Learning}

The use of GNNs for multimodal learning is attractive because of their flexibility to model interactions both within and across different data types. However, data fusion through graph learning requires the construction of network topology and the application of inference algorithms over graphs. We present a methodology that, given a collection of multimodal input data, yields output representations that are used in downstream tasks. We refer to this methodology as \textit{multimodal graph learning} (MGL). MGL can be seen as a blueprint consisting of four learning components that are connected in an end-to-end fashion. In Figure~\ref{fig:mgl_blueprint}a,b, we highlight the difference between a conventional combination of unimodal architectures for treating multimodal data and the suggested all-in-one multimodal architecture.

The first two components of MGL, identifying entities and uncovering topology, can be grouped as the structure learning (SL) phase (Figure~\ref{fig:mgl_blueprint}c) :

\xhdr{Component 1: Identifying Entities}
The first component identifies relevant entities in various data modalities and projects them into a shared namespace. For example, in precision medicine, the state of a patient might be described by matched pathology slides and clinical notes, giving rise to patient nodes with the combined image and language information. In another example from computer vision (Figure~\ref{fig:figure_3t}), entity identification entails defining superpixels in an image.

\xhdr{Component 2: Uncovering Topology}
With the entities of our problem defined, the second component discovers the interactions and interaction types among the nodes across the modalities. Interactions are often explicitly provided, so the graph is given, and this component is responsible for the combination of the already existing graph  structure with the rest of modalities (\eg, in Figure~\ref{fig:figure_5t}c, the Uncovering Topology component corresponds to combining protein surface information with the protein structure itself). When the data does not have an \textit{a priori} network structure, the uncovering topology component explores possible adjacency matrices based on explicit (\eg, spatial and visual characteristics) or implicit (\eg, similarities in representations) features. For the latter case, examples from the natural language processing field consider the construction of graphs from text input that express relations among words (Figure~\ref{fig:figure_4}b).

After graphs are specified or adaptively optimized (SL phase in MGL; Figure~\ref{fig:mgl_blueprint}c), various strategies can be used to learn on the graphs. The last two MGL components, known together as the learning on structure (LoS) phase (Figure~\ref{fig:mgl_blueprint}c), capture these strategies.

\xhdr{Component 3: Propagating Information} 
The third component employs convolutional or message-passing steps to learn node representations based on graph adjacencies~(Supplementary Note~1 for more details on graph convolutions and message passing). In the case of multiple adjacency matrices, methods use independent propagation models or assume a hypergraph formulation that fuses adjacency matrices with a single propagation model. 

\xhdr{Component 4: Mixing Representations} 
The last component transforms learned node-level representations depending on downstream tasks. The propagation models output representations over the nodes that can be combined and mixed depending on the final representation level (\eg, a graph-level or a subgraph-level label). Popular mixing strategies include simple aggregation operators (\eg, summation or averaging) or more sophisticated functions that incorporate neural network architectures.

Figure~\ref{fig:mgl_blueprint}c shows all MGL components, going from multimodal input data to optimized representations used for downstream tasks. Mathematical formulations are in Box~\ref{box:mgl_blueprint} and summaries of multimodal graph learning methods are in Supplementary Note~2. 

\begin{regbox} \textit{\textbf{The blueprint for multimodal graph learning.}} \label{box:mgl_blueprint}
The blueprint for graph-centric multimodal learning has four components.
\begin{enumerate}
    \item \textbf{Identifying Entities}: Information from different sources is combined and projected into a shared namespace. 
    Nodes are identified independently as set elements, and no interactions are given yet.
    Let there be $k$ modalities $\mathcal{C} = \{\mathbf{C}_1, ..., \mathbf{C}_k\}$, where $\mathbf{C}_i$ is an information matrix of $i$-th modality that describes every entity by an information vector. We define $\text{Identify}_i$ module for every modality $i$ as:
    \begin{equation}
        \mathbf{X}_i \leftarrow \text{Identify}_i(\mathbf{C}_i),       
    \end{equation}
    that maps information of all modalities into the same namespace. If $k=1$, we get a reduced unimodal variant of MGL. 
    \item \textbf{Uncovering Topology}: Let there be data modalities $\mathcal{X} = \{\mathbf{X}_1,...,\mathbf{X}_k\}$. We define $\text{Connect}_j$ modules, $j=1,\dots, m$, to specify connections between entities in $\mathcal{X}$ based on $m$ distance measures as:
    \begin{equation}
        \mathbf{A}_j \leftarrow \text{Connect}_j(\mathcal{X}).
    \end{equation}
    If $\mathbf{X}_i$ is already given as an adjacency matrix, the associated $\text{Connect}_j$ modules specify predefined neighborhoods.

    \item \textbf{Propagating Information}: Neural messages are exchanged along edges in the adjacency matrices $\gA = \{\rmA_1,...,\rmA_m\}$ to produce node representations: 
    \begin{equation}
        \rmH \leftarrow \text{Propagate}(\mathcal{X},\mathcal{A}).
    \end{equation}
    When multiple adjacency matrices are given, the $\text{Propagate}$ module can specify multiple independent propagation models (Supplementary Note~1) or operate on a combined adjacency matrix.
    \item \textbf{Mixing Representations}: Representations are mixed and transformed into latent representations optimized for a downstream task:
    \begin{equation}\label{eq:mgl_mix}
        \rmZ \leftarrow \text{Mix}(\mathbf{H},\mathcal{A}).
    \end{equation}
     The mixing module $\text{Mix}$ transforms node representations into final representations of entities $\rmZ$ on which downstream tasks are defined on. Established strategies to mix representations include aggregation operators, such as summation~\cite{xu2018how}, averaging~\cite{hamilton_graphsage}, multi-hop aggregation~\cite{jumping}, and methods using adjacency information $\gA$.
\end{enumerate}
\end{regbox}


\section{Multimodal Graph Learning for Images}\label{sec:images}

 Image-intensive graphs (IIGs) are multimodal graphs where nodes represent visual features and edges represent spatial connections between image features. Structure image learning entails creating IIGs to encode geometric priors relevant to images, such as  translational invariance and scale separation~\cite{bronstein2021geometric}. Translational invariance describes how the output of a CNN must not change depending on shifts in the input image and is achieved by convolutional filters with shared weights. In contrast, scale separation specifies how to decompose long-range interactions between features across scales, focusing on localized interactions that can be propagated to coarser scales. For example, pooling layers follow convolution layers in CNNs to achieve scale separation  \cite{bronstein2021geometric}. In addition, GNNs can model long-range dependencies of arbitrary shape that are important for image-related tasks \cite{chen2018graphbased} such as image segmentation \cite{varga2021fast, liu2021scgnet}, image restoration \cite{zhou2020cross, mou2021restore}, or human object interaction \cite{qi2018learning, wang2020contextual}. 
 
\begin{figure}[t]
    \centering
    \includegraphics[width=\textwidth]{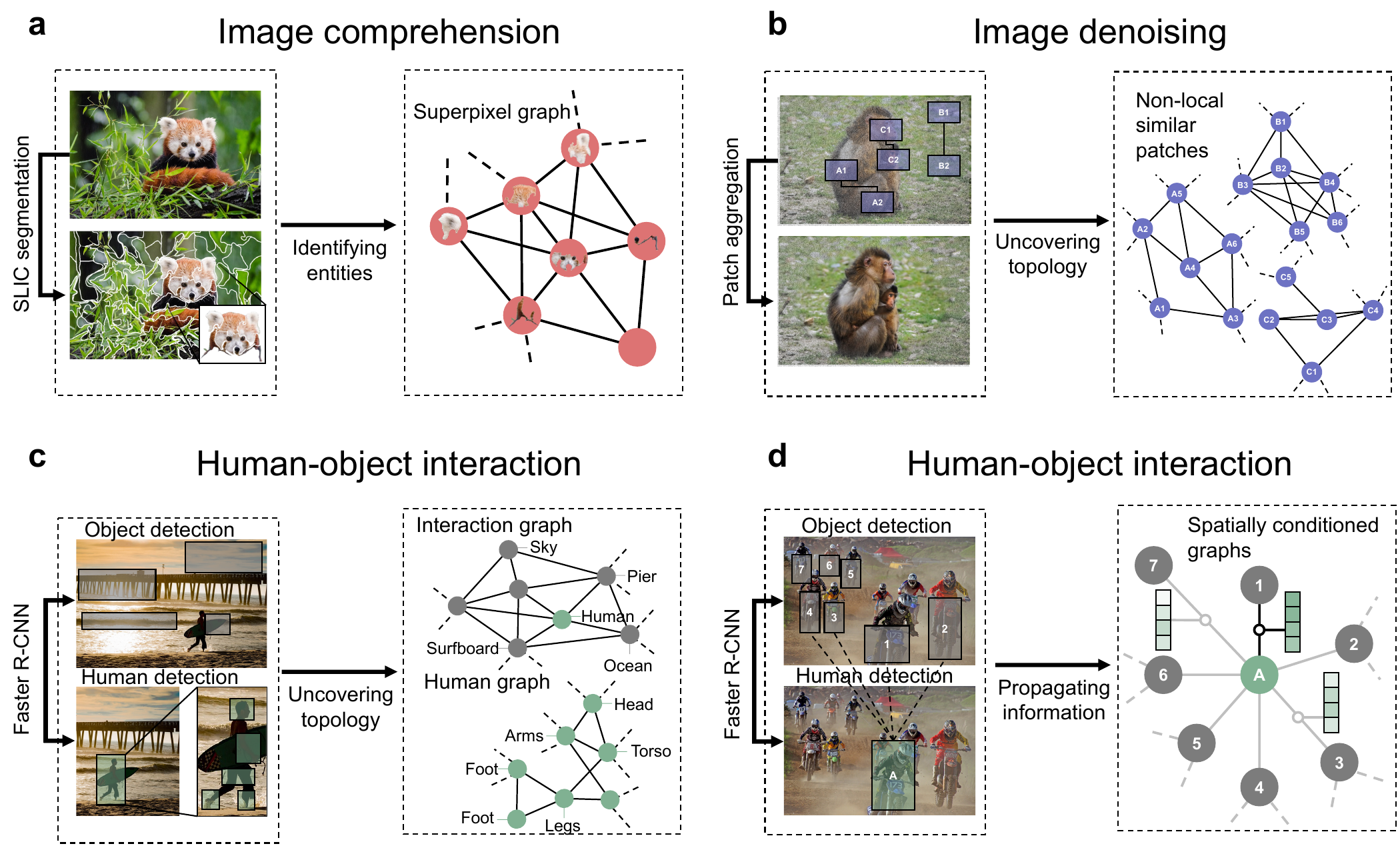}
    \caption{\textbf{Application of multimodal graph learning blueprint to images}. \textbf{a,} Modality identification for image comprehension where nodes represent aggregated regions of interest, or superpixels, generated by the SLIC segmentation algorithm. \textbf{b,} Topology uncovering for image denoising where image patches (nodes) are connected to other non-local similar patches. \textbf{c,} Topology uncovering in human-object interaction where two graphs are created. A human-centric graph maps body parts to their anatomical neighbors, and an interaction connects body parts relative to the distance to other objects in the image. \textbf{d,} Information propagation in human-object interaction where spatially conditioned graphs modify message passing to incorporate edge features that enforce the relative direction of objects in an image \cite{zhang2021spatially}. } 

    \label{fig:figure_3t}
\end{figure}

\xhdrd{Visual Comprehension}
Visual comprehension remains a cornerstone of visual analyses, where multimodal graph learning has proven helpful in classifying, segmenting, and enhancing images. Image classification identifies the set of object categories present in an image \cite{avelar2020superpixel}. In contrast, image segmentation divides an image into segments and assigns each segment into a category~\cite{lu2020graphfcn,varga2021fast,liu2021scgnet}. Finally, image restoration and denoising transform low-quality images into high-resolution counterparts~\cite{kim2016deeplyrecursive}. The information required for these tasks lies in objects, segments, and image patches, as well as in the long-range context surrounding them~\cite{lu2020graphfcn}. 

IIG construction (corresponding to MGL Components 1 and 2) begins with a segmentation algorithm such as simple linear iterative clustering (SLIC)~\cite{SLIC} to identify meaningful regions~\cite{zeng2020semi, wan2019multiscale, varga2021fast} (Figure \ref{fig:figure_3t}a). These regions define nodes used to extract feature maps and summary visual features for each region \cite{liu2021scgnet,lu2020graphfcn}, whose attributes are initialized from CNNs like FCN-16 \cite{long2015fully} or VGG19 \cite{simonyan2015deep}. Moreover, the nodes are connected to their $k$ nearest neighbors in the CNN learned feature space~\cite{zeng2020semi, liu2021scgnet, zhou2020cross, mou2021restore} (Figure \ref{fig:figure_3t}b), to spatially adjacent regions \cite{avelar2020superpixel, knyazev2019image, varga2021fast, wan2019multiscale}, or to an arbitrary number of neighbors based on a previously defined similarity threshold between nodes~\cite{mou2021restore, wan2019multiscale}.

Once the SL phase of MGL is completed, propagation models (MGL Component 3) based on graph convolutions~\cite{lu2020graphfcn, knyazev2019image, wan2019multiscale,liu2021scgnet} and graph attention~\cite{vel2018graph} (GAT) are used to weigh node neighbors in the graph based on learned attention scores \cite{avelar2020superpixel, mou2021restore}. In addition, methods such as graph denoiser networks (GCDNs)~\cite{valsesia2019deep}, internal graph neural networks (IGNNs)~\cite{zhou2020cross}, and residualGCNs \cite{residualGCN, varga2021fast} consider edge similarities to indicate the relative distance between image regions.

\xhdrd{Visual Reasoning}

Visual reasoning goes beyond recognizing visual elements by asking questions about the relationships between entities in images. These relationships can involve humans and objects as in human-object interaction~\cite{qi2018learning} (HOI) or, more broadly, visual, semantic, and numeric entities as in visual question answering~\cite{biten2019scene, singh2019vqa, gsmn} (VQA). 

In HOI, the MGL methods identify two entities, human body parts (\eg, hands, face, etc.) and objects (\eg, surfboard, bike, etc.) \cite{qi2018learning, zhang2021spatially}, that interact in fully connected \cite{qi2018learning, wang2020contextual}, bipartite \cite{zhang2021spatially, ulutan2020vsgnet}, or partially connected topologies \cite{gao2020drg, zhou2019relation}. MGL methods for VQA construct a new topology \cite{gao2020multimodal} that spans interconnected visual, semantic, and numeric graphs. Entities represent visual objects identified by an extractor, such as Faster R-CNN \cite{ren2016faster}, scene text identified by optical character recognition, and number-type texts. Interactions between these entities are defined based on spatial localization: entities occurring near each other are connected by edges.

To learn about these structures (MGL Component 3), methods distinguish between propagating information between entities of the same type and entities of different types. In HOI, knowledge about entities of the same kind (\ie, intra-class neural messages) is exchanged by following edges and applying transformations defined by a GAT \cite{vel2018graph}, which weighs neural messages by the similarity of latent vectors of nodes. In contrast, information between different entities (\ie, inter-class neural messages) is propagated using a GPNN \cite{qi2018learning} where the weights are adaptively learned \cite{wang2020contextual}. Models can have multiple channels that reason over entities of the same class and share information across classes. For example, in HOI, relation parsing neural networks \cite{zhou2019relation} use a two-channel model where human and object-centric message passing is performed before mixing these representations for the final prediction (Figure \ref{fig:figure_3t}c). The same occurs in VQA, where visual, semantic, and numeric channels perform independent message passing before sharing information via visual-semantic aggregation and semantic-numeric aggregation \cite{gao2020multimodal, wu2020ginet}. Other neural architectures can serve as drop-in replacements to graph-based channels~\cite{ulutan2020vsgnet,gao2020drg}.


\section{Multimodal Graph Learning for Language}\label{sec:language}

With the ability to generate contextual language embeddings, language models have broadly reshaped analyses of natural language~\cite{liu2020survey}. However, beyond words, structure in language exists at the level of sentences (syntax trees, dependency parsing), paragraphs (sentence-to-sentence relations), and documents (paragraph-to-paragraph links) \cite{wu2021graph}. Transformers, a prevailing class of language models~\cite{vaswani2017attention}, can capture such structure but have strict computational and data requirements. MGL methods mitigate these issues by infusing language structure into models. Specifically, these methods rely on language-intensive graphs (LIGs), explicit or implicit graphs where nodes represent semantic features linked by language dependencies. 

\xhdrd{Creating Language-Intensive Graphs}

\begin{figure}[t]
    \centering
    \includegraphics[width=\textwidth]{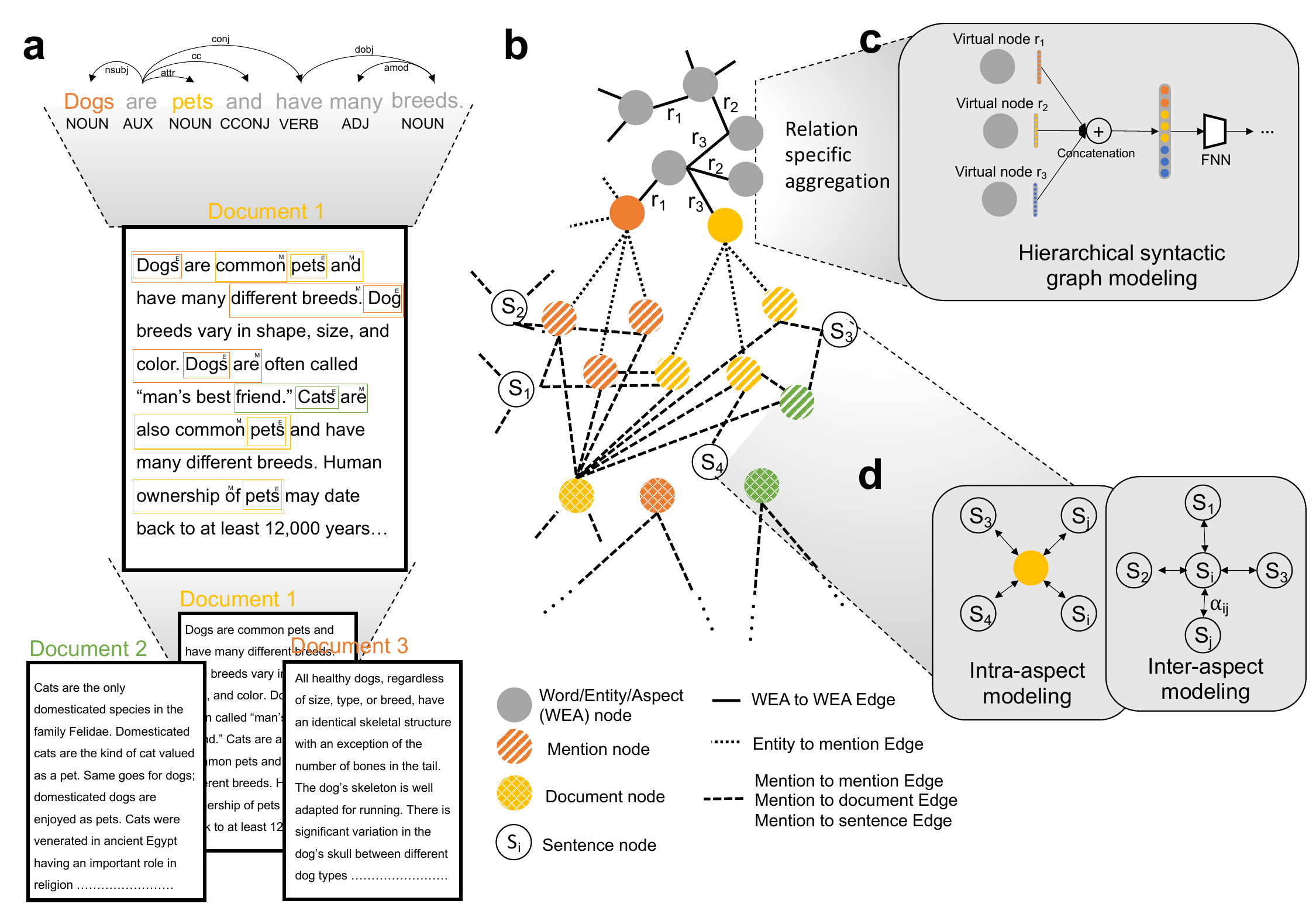}
    \caption{\textbf{Application of multimodal graph learning blueprint to language}. \textbf{a,} The different levels of context in text inputs from sentences to documents and the individual units identified at each context level. This is an example of modality identification's first component of the MGL blueprint. \textbf{b,} The simplified construction of a language-intensive graph from text input, an application of the topology uncovering component of the MGL blueprint. \textbf{c,} and \textbf{d,} visualize examples of learning on LIGs for aspect-based sentiment analysis (ABSA), which aims to assign a sentiment (positive, negative, or neutral) to a sentence with regards to a given aspect. By grouping by relation type from within a sentence (shown in \textbf{c}) or modeling relations between sentences and aspects (shown in \textbf{d}), these methods integrate inductive biases relevant to ABSA and innovate in MGL's third component, information propagation.}

    \label{fig:figure_4}
\end{figure}

At the highest level, a language dataset can be seen as a corpus of documents, then a single document, a group of sentences, a group of mentions, a group of entities, and finally, single words (Figure \ref{fig:figure_4}a). Multimodal graph learning can consider these different levels of contextual information by constructing LIGs. The choice of context to include and how to create a LIG to represent this context is task specific. We describe these steps for text classification and relation extraction as these tasks underlie most language analyses. 

In text classification, the model is asked to assign a label to a span of text \cite{li_heterogeneous_2021} based on the usage and meaning of words (tokens). Graph structure involving words is given by the relative position of words in a document \cite{huang2019text, li_heterogeneous_2021} or document cooccurrence~\cite{zhang2020document}. Relation extraction seeks to identify relations between words in a text, a capability important for other language tasks, such as question answering, summarization, and knowledge graph reasoning \cite{pan2021mentioncentered,zhu_graph_2019}. To capture sentence meaning, the structure among word entities is based on the underlying dependency tree~\cite{guo2020attention}. Beyond words, other entities are included to capture cross-sentence topology \cite{pan2021mentioncentered, zeng2020double} (Figure \ref{fig:figure_4}a-b).

\xhdrd{Learning on Language-Intensive Graphs}

Once a LIG is constructed, a model must be designed to learn on the LIG while incorporating inductive biases relevant to the language task. We illustrate strategies for learning on LIGs using aspect-based sentiment analysis (ABSA) as a downstream language task~\cite{chen-etal-2020-aspect}. ABSA assigns a sentiment (positive, negative) of a text to a word/words or an aspect \cite{chen-etal-2020-aspect}. Models must reason over syntactic structure and long-range relations between aspects and other words in the text to perform ABSA \cite{zhang_aspect-based_2019, zhang-qian-2020-convolution}. To propagate information between distant words, aspect-specific GNNs mask non-aspect words in LIGs for long-range message passing~\cite{zhang_aspect-based_2019}. They also gate or perform element-wise multiplication between latent representations of query and aspect words \cite{veyseh2020improving}. To include information about the syntactic structure, GNNs distinguish between the different types of relations in the dependency tree via type-specific message passing \cite{zhang_aspect-based_2019, zhang-qian-2020-convolution, veyseh2020improving} (Figure \ref{fig:figure_4}c).

The sentiment of neighboring or similar sentences is essential to determine the aspect-based sentiment of the document \cite{chen-etal-2020-aspect}. Cooperative graph attention networks (CoGAN) incorporate this via the cooperation between two graph-based modeling blocks: the inter- and intra-aspect modeling blocks (Figure \ref{fig:figure_4}d) \cite{chen-etal-2020-aspect}. These blocks capture the relation of sentences to other sentences with the same aspect (intra-aspect) and to neighboring sentences in the document that contain different aspects (inter-aspect). The outputs of the intra- and inter-aspect blocks are mixed in an interaction block, passing through a series of hidden layers. Finally, the intermediate representations between each hidden layer are fused via learned attention weights to create a final sentence representation (MGL Component 4).


\section{Multimodal Graph Learning in Natural Sciences }\label{sec:knowledge}

In addition to computer vision and language modeling, graphs are increasingly employed in the natural sciences. We call these graphs knowledge-intensive graphs (KIGs) as they incorporate inductive biases relevant to a specific task or encode scientific knowledge in their structure.

\xhdrd{Multimodal Graph Learning in Physics}

In particle physics, GNNs have been used to identify progenitor particles causing particle jets, sprays of particles that fly out from high-energy particle collisions~\cite{shlomi2021particle}. In these graphs, nodes are particles connected to their $k$-nearest neighbors. After rounds of message passing, aggregated node representations are used to identify progenitor particles~\cite{Henrion2017NeuralMP, qasim2019particle, Mikuni2020ABCNet,ju2020graph}. 

\begin{figure}[t]
    \centering
    \includegraphics[width=\textwidth]{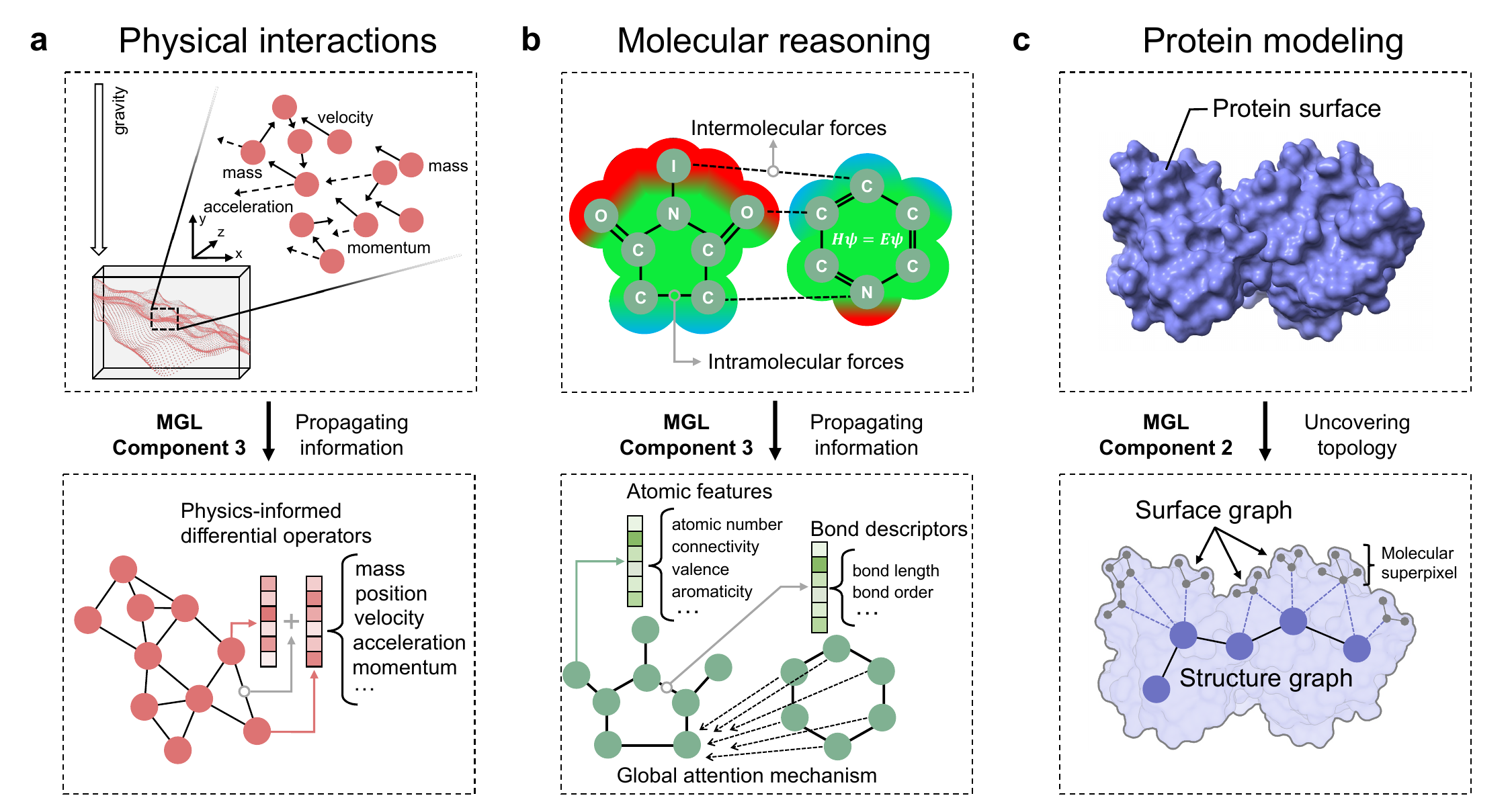}
    \caption{\textbf{Applications of multimodal graph learning to natural sciences}. \textbf{a,} Information propagation in physical interactions where physics-informed neural message passing is used to update the states of particles in a system due to inter-particle interactions and other forces. \textbf{b,} Information propagation in molecular reasoning where a global attention mechanism is used to model the potential interaction between atoms in two molecules to predict whether two molecules will react. \textbf{c,} Topology uncovering in protein modeling where a multiscale graph representation is used to integrate primary, secondary, and tertiary structures of a protein with higher-level protein motifs summarized in molecular superpixels to represent a protein \cite{somnath2021multiscale}. This robust topology provides a better prediction on tasks such as protein-ligand binding affinity prediction.
    } 
    \label{fig:figure_5t}
\end{figure}

Physics-informed GNNs have emerged as a promising approach for simulating physical systems governed by multiscale processes for which conventional methods fail~\cite{shukla2022scalable}. A typical goal is to discover hidden physics from available experimental data. GNNs are trained from available experimental data and information obtained by employing the physical laws and are then evaluated at points in the space-time domain. Such physics-informed architectures integrate multimodal data with mathematical models. For example, GNNs can express differential operators of the underlying dynamics as functions on nodes and edges~\cite{seo2019differentiable}. 
GNNs can also represent physical interactions between objects, such as particles in a fluid \cite{sanchezgonzalez2020learning}, joints in a robot \cite{sanchezgonzalez2018graph}, and points in a power grid \cite{li2021physics}. Initial node representations describe the initial state of these particles and global constants like gravity~\cite{sanchezgonzalez2020learning} with edges indicating relative particle velocity \cite{sanchezgonzalez2018graph}. Message passing updates edge representations first to calculate the effect of relative forces in the system. It then uses the updated edge representation to update node representations and calculate the new state of particles as a result of the forces~\cite{battaglia2018relational} (Figure \ref{fig:figure_5t}a). This message-passing strategy advances the MGL's third component (Section \ref{sec:multimodal}) and has also been employed to solve combinatorial algorithms (Bellman-Ford and Prim's algorithms)~\cite{velickovic2020neural,schuetz2021combinatorial} and chip floorplanning to design the physical layout of computer chips~\cite{mirhoseini2020chip}.

\xhdrd{Multimodal Graph Learning in Chemistry}

In chemistry, MGL methods can predict intra- and inter-molecular properties from the primary molecular structure by performing message passing on molecular graphs of atoms linked by bonds \cite{pmlr-v70-gilmer17a, klicpera2020directional, jorgensen2018neural, klicpera2021directional,liu2021fast, john2020orgo}. Present efforts incorporate 3D spatial molecular information in addition to 2D molecular details. When this information is unavailable, the MGL methods~\cite{klicpera2020directional, klicpera2021directional, liu2021fast} consider stereo-chemistry to aggregate neural messages \cite{pattanaik2020message} and model molecules as sets of chemical substructures in addition to granular atom representations~\cite{fey2020hierarchical}. 

Stereoisomers are molecules with the same graph connectivity but different spatial arrangements~\cite{pattanaik2020message}. Aggregation functions in molecular graphs aggregate the same regardless of the orientation of atoms in three-dimensional space. This can lead to poor performance, as stereoisomers can have different properties~\cite{ariens1986chiral}. To mitigate this issue, permutation (PERM) and permutation-concatenation (PERM-CAT) aggregation~\cite{pattanaik2020message} update every atom in a chiral group via a weighted sum of every permutation of its respective chiral group. Though the identity of the neighbors is the same in every permutation, the spatial arrangement varies. By weighing each permutation, PERM and PERM-CAT encode this inductive bias by modifying how information is propagated in the underlying graph (MGL Component 3). 

Moreover, MGL can help identify chemical products produced by molecules through reactions~\cite{guan2021selectivity,coley2019react,struble2020aromatic,stuyver2021quantum}. For example, to predict whether two molecules react, QM-GNN~\cite{guan2021selectivity}, a quantum chemistry-augmented GNN represents each reactant by its molecular graph with chemistry-informed initial representations for every atom and bond. After rounds of message passing, the atom representations are updated through a global attention mechanism (Figure \ref{fig:figure_5t}b). The attention mechanism uncovers a novel topology where atoms can interact with atoms on other molecules. It incorporates a principle from chemistry that intermolecular interactions between particles inform reactivity. The final representations are combined with descriptors, such as atomic charges and bond lengths, and used for prediction. Such an approach integrates structural knowledge about molecules in a GNN with relevant chemistry knowledge, allowing for accurate prediction on small training datasets~\cite{guan2021selectivity}. The inclusion of domain knowledge by fusing GNN outputs illustrates the $\text{Mix}$ module in MGL (Section \ref{sec:multimodal}, Box \ref{box:mgl_blueprint}). Graph learning on molecules created new opportunities for virtual drug screening~\cite{stokes2020antibiotic}, molecule generation and design~\cite{fu2021differentiable,Mercado2021design,walters2021}, and drug target identification~\cite{torng2019DTI,moon2021pignet}.

\xhdrd{Multimodal Graph Learning in Biology}

Beyond individual molecules, MGL can help understand the properties of complex structures across multiple scales, the most pertinent of these structures being proteins. At the primary amino acid sequence scale, the hallmark task predicts the 3D structure from the amino acid sequence. AlphaFold constructs a KIG where nodes are amino acids with representations derived from sequence homology \cite{jumper2021alphafold}. To propagate information in this KIG, AlphaFold introduces a triangle multiplicative update and triangle self-attention update. These triangle modifications integrate the inductive bias that learned representations must abide by the triangle inequality on distances to represent 3D structures. Multimodal graph learning, among other innovations, enabled AlphaFold to predict 3D protein structure from amino acid sequence \cite{jumper2021alphafold}.

Beyond 3D structure, molecular protein surfaces mediate critical roles in cellular function and disease, and thus modeling geometric and physical protein properties is essential~\cite{Greener2022,gainza2020, sanner1996reduced}. For example, MaSIF~\cite{gainza2020} trains a GNN on molecular surfaces described as multimodal graphs to predict protein interactions.
The initial representation of the nodes is based on geometric and chemical features. Next, Gaussian kernels are defined on every node to propagate information, encoding complex geometric shapes of molecular surfaces and extending the notion of a convolution. The final representations can be used to predict protein-protein interactions~\cite{gainza2020}, structural configurations of protein complexes~\cite{sverrisson2021CVPR}, and protein-ligand binding~\cite{somnath2021multiscale}.


\section{Outlook}

Multimodal graph learning is an emerging field with applications across natural sciences, vision and language domains. We anticipate the growth in MGL be driven by fully multimodal graph architectures and new uses in the natural sciences and medicine. We also outline applications to understand when MGL is valuable or unhelpful and needs improvements to resolve challenges represented by multimodal inductive biases or a lack of explicit graphs.

\xhdrd{Fully Multimodal Graph Architectures} 

Prevailing approaches use domain-specialized architectures tailored to each data modality. However, advances in general-purpose architectures provide an expressive strategy to consider dependencies between modalities irrespective of whether they are given as images, language sequences, graphs, or tabular datasets. Moreover, the MGL blueprint supports more complex graph structures, such as hypergraphs~\cite{feng2019hypergraph,hypergraphs2021,hypergraphs_edge_rep} and heterogeneous graphs~\cite{hetero_gnn,Chandak2022.05.01.489928}. 

The blueprint can also pave the way for novel uses of graph-centric multimodal learning. For example, knowledge distillation (KD) aims to transfer knowledge from a teacher model to a smaller student model in a way that preserves performance while using fewer resources. Knowledge-intensive graphs~\cite{lee2019graphbased,zhou2021distilling,sun2021collaborative} can be used to design more efficient KD loss functions~\cite{park2019relational,liu2019distillation}. In another example, visible neural networks specify the architecture such that nodes correspond to concepts (\eg, molecules, pathways) at different scales of the cellular system, ranging from small complexes to extensive signal pathways~\cite{michael2018visible,ma2018using}, connected based on biological relationships, used in forward- and back-propagation. By incorporating such inductive biases, models can be trained in a data-efficient manner as they do not have to invent relevant fundamental principles but can know these from the start and thus need fewer data for training. Harmonizing algorithm design with domain knowledge can also improve model interpretability.

\xhdrd{Algorithmic Improvements to Resolve Multimodal Challenges}

Existing methods are limited in areas without prior knowledge or relational structure. For example, in tasks such as chemical reaction prediction \cite{guan2021selectivity}, progenitor particle classification \cite{shlomi2021particle}, physical interaction simulation \cite{sanchezgonzalez2020learning}, and protein-ligand modeling \cite{gainza2020}, interactions relevant for the task are not a priori given, meaning that the methods must automatically capture novel, unspecified, and relevant interactions. Some applications use node feature similarity to dynamically construct local adjacencies after each layer to discover new interactions \cite{shlomi2021particle}. However, this cannot capture novel interactions among distant nodes since information is only passed among closely connected nodes in message passing. Methods address this limitation by incorporating attention layers with induced sparsity to discover these interactions \cite{guan2021selectivity}. In applications  without strong relational structure, such as molecular property prediction \cite{klicpera2021directional,liu2021fast, john2020orgo}, particle classification~\cite{shlomi2021particle}, and text classification~\cite{li_heterogeneous_2021}, node features often have more predictive value than any encoded structure. As a result, other methods have been shown to lead to better performance than graph-based methods \cite{Borisov2021Tabular, Jiang2021chemistry}.

\xhdrd{Groundbreaking Applications in Natural Sciences and Medicine}

Using deep learning in natural sciences revealed the power of graph representations for modeling small and large molecular structures. Combining different types of data can create bridges between the molecular and organism levels for modeling physical, chemical, or biological phenomena on a large scale. Recent knowledge graph applications have been introduced to enable precision medicine and make predictions across genomic, pharmaceutical, and clinical applications~\cite{Chandak2022.05.01.489928, NICHOLSON20201414}. Multi-scale learning systems are becoming valuable tools for protein structure prediction~\cite{jumper2021alphafold}, protein property prediction~\cite{somnath2021multiscale},  and biomolecular interaction modeling~\cite{pan2021mentioncentered}. These methods can incorporate mathematical statements of physical relationships, knowledge graphs, prior distributions, and constraints by modeling predefined graph structures or modifying message-passing algorithms. When such information exists, multimodal learning can enhance image denoising~\cite{kim2016deeplyrecursive}, image restoration~\cite{kim2016deeplyrecursive}, and human-object interaction~\cite{qi2018learning} in vision systems.

\clearpage

\paragraph{Data and code availability}

We summarize multimodal graph learning (MGL) methods and provide a continually updated summary at \url{https://yashaektefaie.github.io/mgl}. We host a live table where future MGL methods will be added as a resource to the community. 

\paragraph{Acknowledgements}

Y.E., G.D., and M.Z. gratefully acknowledge the support of US Air Force Contract No. FA8702-15-D-0001, and awards from Harvard Data Science Initiative, Amazon Research, Bayer Early Excellence in Science, AstraZeneca Research, and Roche Alliance with Distinguished Scientists. Y.E. is supported by grant T32 HG002295 from the National Human Genome Research Institute and the NSDEG fellowship. G.D. is supported by the Harvard Data Science Initiative Postdoctoral Fellowship. Any opinions, findings, conclusions or recommendations expressed in this material are those of the authors and do not necessarily reflect the views of the funders.
 
\paragraph{Competing interests} 

The authors declare no competing interests.

\clearpage


\bibliographystyle{naturemag.bst}

\clearpage

\pagenumbering{arabic}
\renewcommand*{\thepage}{S\arabic{page}}
\appendix

\section*{Supplementary Note 1: Overview of Graph Neural Networks}

Graph representation learning has been at the center of an accelerating research area due to the prominence of the so-called graph neural network (GNN) models. GNNs have become very popular, mainly because of their wide applicability and their flexible framework to define simple or more complex information propagation processes. In most applications, however, the graph learning models require predefined graphs to apply diffusion operators to them.

\subsection*{Graphs and Attributes}

In the standard formulation of learning methods on graphs, we let a graph $\mathcal{G}$ be defined as a tuple of two sets, a set of nodes $V$ and a set of edges $\mathcal{E}: \mathcal{G} = (V,\mathcal{E})$. Edges correspond to the interactions that nodes share with each other and, thus, we consider them \textit{adjacent}: $\mathcal{E} = \{(u_i,u_j) | u_i \in V,\ u_j\in V,\ \text{there is an edge from $u_i$ to $u_j$} \}$. In order to encode the adjacency property, various matrices are suggested depending on the model and the task. Most popular ones are the \textit{adjacency matrix} $\rmA \in \{0,1\}^{n \times n}$, where $\mathbf{A}_{ij} = 1$ if and only if $(i,j) \in \mathcal{E}$, the \textit{Laplacian matrix} $\mathbf{L} = \mathbf{D} - \mathbf{A}$, where $\mathbf{D} = Diag( \mathbf{A} \mathbf{1}_n),$ is the \textit{degree matrix}, as well as variants of them.

Graph learning consists of extracting knowledge from a graph or its components. For this knowledge extraction, we need to model explicitly the contextual information that the nodes, the edges, or the whole graph can carry. For example, in a molecular network, the nodes and the edges might carry information about the atom type and the bond type respectively~\cite{moleculenet}. In order to encode the context, the \textit{attribute matrices} are being defined: $\mathbf{X}_v \in \mathbb{R}^{n \times l_v}$ and $\mathbf{X}_e \in \mathbb{R}^{m \times l_e}$ for the information over the nodes and the edges respectively.  

\subsection*{Tasks and Representation Levels}

In graph learning tasks, the main objective is the label prediction of either nodes, edges, or a whole graph. This creates a categorization among the models and tasks based on the label type:
\begin{itemize}
    \item \textit{Node-level representations}: Node representations ($\rmH \in \mathbb{R}^{n\times d_v}$) are needed when the objective is a node-level prediction task, such as node classification and node regression~\cite{kipf_gcn, hamilton_graphsage}. Given a subset of labeled nodes, the model makes predictions for unlabeled nodes in the graph.
    \item \textit{Edge-level representations}: The domain of edge-level representations spans $V^2$, that describes the possible pairs among graph nodes ($\rmH \in \mathbb{R}^{m\times d_e}$). Edge classification~\cite{Kim_2019_CVPR} and link prediction~\cite{seal_gnn} are two tasks that edge-level representations that are often utilized. 
    \item \textit{Graph-level representations}: Here, a single representation describes a graph ($\rmH \in \mathbb{R}^{d_g}$). Such a representation is required for graph-level tasks, such as graph classification~\cite{xu2018how, pmlr-v139-beani21a} and graph regression~\cite{pmlr-v70-gilmer17a,bodnar2021weisfeiler}, where the goal is the prediction of a graph label.
    \item \textit{Substructure-level representations}: A graph substructure or \textit{subgraph} can be defined as a tuple $(V_s, \gE_s)$, where $V_s \subseteq V$ and $\gE_s = \{(u_i,u_j) | u_i \in V_s, u_j \in V_s, (u_i,u_j) \in \gE \}$. Similarly, we can define classification and regression tasks upon the nodes or the edges of subgraphs~\cite{alsentzer2020subgraph}. Node-, edge- and graph-level representations can be seen as instances of substructure-level representations.
\end{itemize}

\subsection*{Graph Neural Architectures}

GNNs can be specified based on spectral forms, recurrent neural schemes, and neural message-passing strategies. At the heart of GNNs is the concept of \textit{neural message-passing}~\cite{pmlr-v70-gilmer17a}. That is the process of propagating information between a node and its neighborhood in a differentiable manner. Assigning a parameterization over the node attributes and using the information propagation model, we learn node representations using gradient-based optimization. Depending on the user-end task, the learned node representations can then be transformed into an appropriate form.

Despite the wide variety of models, a GNN layer is characterized by the following three computational steps: 1) First, \textit{neighborhood aggregation} is a differentiable function responsible for generating a representation for a node's neighborhood: 
\begin{equation} \label{eq:sec_2_agg}
    h_u^{\text{agg}} = \text{\textbf{Agg}}(\{h_v | v \in \mathcal{N}_u\}),
\end{equation}

where $\mathcal{N}_u$ is the neighborhood of node $u$. Although $\mathcal{N}_u$ can be defined in many ways, in most GNN architectures~\cite{scarselli_gnn, kipf_gcn}, it corresponds to the nodes that are adjacent to the examined node (i.e., the $1$-hop neighborhood). The choice of the aggregation function can output very different architectures and can be classified into two main categories, the convolutional and the message-passing models, as discussed in Box~S\ref{box:gnns_variants}. 2) Second, \textit{node update} is a differentiable function that learns node representations by combining the current state of the nodes with the aggregated representation of their neighborhoods: 
\begin{equation}  \label{eq:sec_2_upd}
    h_u' = \text{\textbf{Update}}(h_u,h_u^{\text{agg}}).
\end{equation} 
3) Finally, \textit{representation transformation} denotes the application of a parametric or non-parametric function that transforms the learned node representations into final embeddings that correspond to the target prediction labels as: 
\begin{equation}  \label{eq:sec_2_trans}
    h_\text{out} = \text{\textbf{Transform}}(\{h_u' | u \in V\}).
\end{equation}

Specifically, for edge-level tasks, such as link prediction and edge labeling, non-parametric operators can be used (\eg concatenation, summation, or averaging of the embeddings of node pairs~\cite{seal_gnn, Kim_2019_CVPR}). For graph-level tasks, a parametric or non-parametric function that mixes node representations is referred to as \textbf{Readout}. Examples of the \textbf{Readout} functions include summation (\eg in graph isomorphism networks~\cite{xu2018how}) and a Set2Set function~\cite{set2set} (\eg in principal neighborhood aggregation~\cite{pna2020}). 

\begin{suppregbox} \textit{\textbf{Types of aggregation}} \label{box:gnns_variants} The choice of the $\textbf{Agg}$ function is crucial for a graph neural network, as it defines the way that neighborhood information is processed. Based on this choice, we meet two large model categories.

\xhdr{Graph convolutions}
The first category is based on the connection of propagation models with graph signal processing~\cite{defferrard2016}.
In particular, assuming a filter $g\in \mathbb{R}^n$ that expresses the behavior of graph nodes, the information propagation can be expressed through a graph convolution: $x \ast g = UgU^Tx$. The matrix $U\in \mathbb{R}^{n\times n}$ is the set of eigenvectors of the Laplacian matrix $L$ expressed as a transformed graph adjacency information. 
Simplifying the graph convolution into an aggregation scheme, we obtain the following:
\begin{equation}
    \label{eq:gnn_convolutional}
    \text{\textbf{Agg}}(\{h_v | v \in \mathcal{N}_u\}) =  \phi(\{\Theta_{uv}\psi(x_v)\, |\, v \in \gN_u\})),
\end{equation}
where $\psi$ is a learnable function such as a multilayer perceptron (MLP) and $\phi$ is an aggregation operator, depending on the choice of filter $g$. For example, in both GCN~\cite{kipf_gcn} and SGC~\cite{sgc2019} models, $\psi$ is the identity function, $\Theta_{uv} = 1 \, \forall \, (u,v)\in \gE$ and the $\phi$ is the average function. Other typical examples of this category are the Chebnet~\cite{defferrard2016} and the Caleynet~\cite{levie2018cayleynets} models, where the choices of $\mathbf{\Theta},\phi$, and $\psi$ are functions of the used graph filters.

\xhdr{Neural message-passing} 
In message passing, aggregation is a process of encoding messages for each node based on its neighbors. The difference with the convolutional variant is that the edge $(u,v) $'s importance is parametrized compared to a non-parametric constant $\Theta_{uv}$. Thus, aggregation is formulated as:
\begin{equation}
    \label{eq:gnn_mp}
    \text{\textbf{Agg}}(\{h_v | v \in \mathcal{N}_u\}) =  \phi(\{\psi(x_u,x_v)\, |\, v \in \gN_u\})),
\end{equation}
where $\psi$ is, similarly, a learnable function (\eg an MLP) and $\phi$ is an aggregator operator. For instance, the family of MPNNs~\cite{pmlr-v70-gilmer17a} utilizes a summation operator as $\phi$, and $\psi$ can take various forms from simple concatenation to a trainable neural network. Similarly, the GIN model~\cite{xu2018how} chooses as $\phi$ is the summation operator and as $\psi$ is an MLP. Going beyond the choice of a single function, ~\cite{pna2020} combines multiple aggregators with degree scalers. 

In this category, we can also meet abundant models, whose functions $\phi$ and $\psi$ are based on attention mechanisms tailored for graph structures. The GAT~\cite{gat} model was the first approach that defined self-attention coefficients for graphs. Since then, attention-based models in combination with learned positional or spectral encodings have been introduced, such as the SAN~\cite{san2020} and Graph Transformer archictures, e.g. Graphormer~\cite{graphormer}, and GraphGPS~\cite{rampasek2022recipe}.

\end{suppregbox}


\section*{Supplementary Note 2: Existing Methods in the MGL Blueprint}

We show how existing methods can be decomposed under the MGL blueprint. Table~\ref{tab:classification} breaks down existing methods according to the four components of the MGL.

\begin{table}[h]
    \captionsetup{labelformat=empty}
    \caption{\textbf{Tab. S1} Classification of existing methods according to the Multimodal Graph Learning (MGL) blueprint. The four components of MGL are identified for every method. Shown are methods for image-intensive graphs (dark gray), language-intensive graphs (light gray), and knowledge-grounded graphs (white). }
    \label{tab:classification}
  \resizebox{\textwidth}{!}{%

\begin{tabular}{cp{3.9cm}p{3.9cm}p{4.9cm}p{4.5cm}p{3.7cm}}
\toprule

\textbf{Method}                 & \multicolumn{1}{c}{\textbf{Identifying entities}} & \multicolumn{1}{c}{\textbf{Uncovering topology}}                                       & \multicolumn{1}{c}{\textbf{Propagating information}}         & \multicolumn{1}{c}{\textbf{Mixing representations}}            & \multicolumn{1}{c}{\textbf{Application}} \\     \toprule
\rowcolor{Gray1}

FuNet~\cite{miniGCN}                  & Hyperspectral pixels                                              & Radial basis function similarity                                  & miniGCN  (GCN mini-batching)                                            & Concatenation, sum, or product & Hyperspectral image classification                       \\ 
\midrule
\rowcolor{Gray1}

Graph-FCN~\cite{lu2020graphfcn_arxiv}              & Pixels                                                            & Edge weights based on a Gaussian kernel & GCN on weighted edges                                & Graph loss added with fully connected network           & Image semantic segmentation                              \\ \midrule
\rowcolor{Gray1}

GSMN~\cite{gsmn}                   & Images, relations, and attributes                                 & Visual graph for images combined with textual graph                              & Node-level and graph-level matching                           & Similarity function for positive and negative pairs                      & Image-text matching                                      \\
\midrule
\rowcolor{Gray1}
RAG-GAT~\cite{avelar2020superpixel}                   & Super-pixels                                 & Region adjacency graph                              & Graph attention network                           & Sum pooling combined with an MLP                      & Superpixel image classification                                      \\

\midrule\midrule \rowcolor{Gray2}

TextGCN~\cite{textGCN}                & Words and documents                                               & Occurrence edges in text and  corpus       & GCN                                                                          & No mixing, single-channel model                                                & Text classification                                      \\
\midrule \rowcolor{Gray2}
CoGAN~\cite{chen-etal-2020-aspect}                  & Sentences and aspects                                             & Sentences and aspects as nodes                        & Cooperative graph attention & Softmax decoding block                       & Aspect sentiment classification                          \\ \midrule \rowcolor{Gray2}
MCN~\cite{pan2021mentioncentered}                    & Sentences, mentions, and entities                                 & Document-level graph & Relation-aware GCN                                                           & Sigmoid activation on entity pairs                               & Document-level relation extraction                       \\ 
\midrule
\rowcolor{Gray2}

GP-GNN~\cite{zhu_graph_2019}                & Word and position encodings                                               & Generated adjacency Matrix       & Neural message passing                                                                          & Pair-wise product                                                & Relation extraction                                      \\

\midrule\midrule
QM-GNN~\cite{guan2021selectivity}                & Atoms                                                             & Chemical bonds & Weisfeiler-Lehman network and global attention                               & Concatenation with quantum mechanical descriptors           & Regio-selectivity prediction                             \\ \midrule
GNS~\cite{sanchezgonzalez2020learning}                    & particles                                                         & Radial particle connectivity  & Graph network (learned directed message passing)                             & No mixing, single-channel model                                                & Particle-based simulation                                \\ \midrule
MaSIF~\cite{gainza2020}                  & Discretized protein surface mesh                      & Overlapping geodesic radial features  & Gaussian kernels on a local geodesic system          & No mixing, single-channel model                                                & Ligand site prediction and classification                \\ \midrule
MMGL~\cite{mmgl}                   & Patients                                                          & Modality-aware latent graph                                                       & Adaptive GCN                                       & Sub-branch prediction neural network                                           & Disease prediction                                       \\      \bottomrule

\end{tabular}}
\end{table}

\end{document}